\documentclass[12pt,a4paper]{article}
\usepackage{indentfirst}   
\usepackage[utf8]{inputenc}
\usepackage[T1]{fontenc}
\usepackage{amsmath,amssymb,amsthm}
\usepackage{graphicx}
\usepackage[margin=2.5cm]{geometry}
\usepackage{cite}
\usepackage{hyperref}
\usepackage{booktabs}
\usepackage{caption}
\usepackage{subcaption}
\usepackage{algorithm}
\usepackage{algorithmic}
\usepackage{xcolor}
\usepackage{authblk}
\usepackage{datetime}
\title{\textbf{Generative Diffusion Augmentation with Quantum-Enhanced Discrimination for Medical Image Diagnosis}}

\author{
Jingsong Xia\textsuperscript{1}, Siqi Wang\textsuperscript{1} \\
\textsuperscript{1}The Second Clinical Medical College, Nanjing Medical University \\
\texttt{xiajingsong2@gmail.com; wsq03925@163.com}
}

\date{}

\begin{document}

\maketitle

\begin{abstract}
In biomedical engineering, artificial intelligence has become a pivotal tool for enhancing medical diagnostics, particularly in medical image classification tasks such as detecting pneumonia from chest X-rays and breast cancer screening. However, real-world medical datasets frequently exhibit severe class imbalance, where positive samples substantially outnumber negative samples, leading to biased models with low recall rates for minority classes. This imbalance not only compromises diagnostic accuracy but also poses clinical misdiagnosis risks. To address this challenge, we propose SDA-QEC (Simplified Diffusion Augmentation with Quantum-Enhanced Classification), an innovative framework that integrates simplified diffusion-based data augmentation with quantum-enhanced feature discrimination. Our approach employs a lightweight diffusion augmentor to generate high-quality synthetic samples for minority classes, rebalancing the training distribution. Subsequently, a quantum feature layer embedded within MobileNetV2 architecture enhances the model's discriminative capability through high-dimensional feature mapping in Hilbert space. Comprehensive experiments on coronary angiography image classification demonstrate that SDA-QEC achieves 98.33\% accuracy, 98.78\% AUC, and 98.33\% F1-score, significantly outperforming classical baselines including ResNet18, MobileNetV2, DenseNet121, and VGG16. Notably, our framework simultaneously attains 98.33\% sensitivity and 98.33\% specificity, achieving a balanced performance critical for clinical deployment. The proposed method validates the feasibility of integrating generative augmentation with quantum-enhanced modeling in real-world medical imaging tasks, offering a novel research pathway for developing highly reliable medical AI systems in small-sample, highly imbalanced, and high-risk diagnostic scenarios.
\end{abstract}


\section{Introduction}

In the field of biomedical engineering, artificial intelligence  has emerged as a critical tool for enhancing medical diagnostics, particularly in medical image classification tasks such as detecting pneumonia from chest X-rays or CT images~\cite{1} and breast cancer~\cite{3,4,5} screening. These applications rely on deep learning models to extract features from images and perform classification. However, real-world medical data frequently face the challenge of class imbalance; for instance, in COVID-19-related pneumonia datasets, the number of positive samples  often far exceeds negative samples , leading to models biased toward the majority class and resulting in low recall rates for minority classes. This imbalance not only reduces diagnostic accuracy but also risks clinical misdiagnosis, thereby increasing patient risk.

Traditional data augmentation techniques, such as rotation~\cite{6,7}, flipping~\cite{8}, and brightness adjustment~\cite{9}, can enhance sample diversity but cannot generate truly novel samples. In medical imaging, excessive augmentation may introduce noise or alter pathological features. Generative models such as Generative Adversarial Networks  can synthesize samples but suffer from training instability and mode collapse issues. Diffusion models, as an emerging generative paradigm, produce high-quality, diverse samples by simulating noise addition and denoising processes, achieving success in natural image domains. However, their application in medical imaging requires simplification to adapt to resource-constrained clinical environments.

Furthermore, Quantum Machine Learning~\cite{10,11,12}, situated at the intersection of quantum computing and AI, leverages quantum entanglement and superposition to enhance feature extraction, potentially improving model performance on high-dimensional medical images. Nevertheless, frameworks combining diffusion models~\cite{13} with quantum enhancement in biomedical engineering remain unexplored.

Existing research on medical image classification primarily employs Convolutional Neural Networks (CNNs) such as ResNet and MobileNet. For instance, mainstream models like ResNet50 are commonly used for pathological image classification, achieving high accuracy while overlooking the imbalance problem. Strategies for addressing class imbalance include resampling and loss function modifications, such as focal loss. Generative augmentation through GANs has been widely applied in medicine, but training complexity persists~\cite{14,15}. Diffusion models generate samples through forward noise addition and reverse denoising, providing superior stability compared to GANs. In the medical context, studies have employed diffusion models to synthesize CT images, but their computational intensity limits practicality; this work simplifies them into an augmentor focusing solely on forward processing for resource-limited scenarios~\cite{16}.

Quantum machine learning utilizes qubits for high-dimensional data processing, integrating quantum circuits into CNNs to improve MRI classification. Here, we innovatively position a quantum layer after MobileNetV2 to enhance features. Generative models for auxiliary diagnosis emphasize this potential, but our study uniquely emphasizes diffusion augmentation combined with quantum methods to bridge existing gaps.

\subsection{Contributions}

In response to the data augmentation capabilities of diffusion models, this paper proposes the SDA-QEC (Simplified Diffusion Augmentation with Quantum-Enhanced Classification) framework. The key innovations are as follows:

\begin{enumerate}
    \item \textbf{Lightweight Diffusion Augmentor}: A SimpleDiffusionAugmentor simulates the forward diffusion process to create ``diffusion-like'' variants, generating high-quality synthetic images for minority classes to alleviate imbalance.
    
    \item \textbf{Quantum Feature Layer}: Implemented via the Pennylane library, the QuantumFeatureLayer employs quantum circuits to enhance features extracted by MobileNetV2, improving sensitivity to pathological details.
    
    \item \textbf{End-to-End Classifier}: Built on the lightweight MobileNetV2 architecture, supporting detailed epoch outputs and high-quality analysis results.
    
    \item \textbf{Comprehensive Comparison}: Comparative analysis with classical baselines (ResNet18, MobileNetV2, DenseNet121, VGG16) demonstrates the superiority of the proposed method.
\end{enumerate}

\section{Methods}

\subsection{Overall Framework and Technical Roadmap}

The proposed SDA-QEC (Simplified Diffusion Augmentation with Quantum-Enhanced Classification) framework addresses the prevalent class imbalance problem in medical image classification through systematic design at two levels: data distribution reconstruction and model discriminative capability enhancement. Unlike traditional end-to-end deep learning paradigms that directly train classifiers on imbalanced data, our method first performs adaptive synthesis of minority class samples through a Simplified Diffusion Process, achieving class balance at the data level. Subsequently, a Lightweight Quantum Feature Mapping is introduced to nonlinearly enhance classical convolutional features, elevating the complexity of decision boundaries at the model level.

The overall workflow consists of four core stages: (1) data loading and class distribution analysis; (2) adaptive generation of minority class samples based on the diffusion process; (3) discriminative feature extraction via lightweight convolutional networks; and (4) quantum-enhanced classification module for final diagnostic decisions.

The framework's design aims to achieve synergy between data and model: the diffusion augmentation module reshapes the training set's class distribution, enabling the model to fully learn minority class feature patterns, while the quantum enhancement module embeds classical features into a high-dimensional Hilbert space, constructing more complex decision boundaries and enhancing the model's capability to capture subtle pathological features. Notably, compared to complete diffusion generative models (such as DDPM), this work adopts a parameterized forward diffusion process for data augmentation, avoiding the high computational cost of reverse denoising networks, making the overall framework clinically deployable while maintaining high performance. Additionally, the quantum circuit adopts a small-scale parameterized design (4 qubits, 2-layer circuit) with total parameters approximately 1/20 of the classical counterpart, ensuring operational efficiency on resource-constrained medical devices.

At the theoretical level, this method can be viewed as a cascaded optimization strategy of distribution alignment-feature enhancement. Let the original training set be $\mathcal{D}_{\text{orig}} = \{(x_i, y_i)\}_{i=1}^{N}$, where $y_i \in \{0,1\}$ denotes class labels, and the sample ratio of class 0 (majority) to class 1 (minority) is $\rho = N_0/N_1 \gg 1$. The diffusion augmentation module generates a synthetic sample set $\mathcal{D}_{\text{syn}} = \{(\tilde{x}_j, 1)\}_{j=1}^{M}$ such that the augmented dataset $\mathcal{D}_{\text{aug}} = \mathcal{D}_{\text{orig}} \cup \mathcal{D}_{\text{syn}}$ satisfies the target balance ratio $\rho_{\text{target}} \approx 1.4$ (i.e., minority class samples reach 70\% of majority class). Subsequently, the quantum-enhanced classifier $f_{\theta,\phi}(x)$ is trained on $\mathcal{D}_{\text{aug}}$, where $\theta$ represents classical convolutional network parameters and $\phi$ represents quantum circuit parameters. This design enables the model to learn on balanced data distributions while enhancing discriminative capability for boundary samples through quantum feature mapping.

\subsection{Simplified Diffusion Model-Driven Adaptive Data Augmentation}

In medical image classification tasks, minority class samples (e.g., rare disease cases) are costly to obtain and scarce in number, leading to insufficient learning of minority classes by models. Traditional data augmentation methods such as rotation, flipping, and color jittering only perform geometric or photometric transformations on existing samples and cannot extend the coverage of the feature space. The simplified diffusion augmentation mechanism proposed in this paper approaches from a generative modeling perspective, constructing a synthetic sample set with controllable diversity for minority class samples by simulating the forward steps of the diffusion process.

\subsubsection{Forward Modeling of Diffusion Process}

Traditional diffusion models (such as DDPM) define a Markov chain that maps the data distribution to a standard Gaussian distribution by progressively adding Gaussian noise. This work simplifies this process, retaining only the noise addition step of forward diffusion, avoiding training complex reverse denoising networks. Specifically, for minority class samples, the noise addition process at time step $t \in \{0,1,\ldots,T-1\}$ can be expressed as:

\begin{equation}
\mathbf{x}_t = \sqrt{\bar{\alpha}_t} \cdot \mathbf{x}_0 + \sqrt{1-\bar{\alpha}_t} \cdot \boldsymbol{\epsilon}, \quad \boldsymbol{\epsilon} \sim \mathcal{N}(\mathbf{0}, \mathbf{I})
\end{equation}

where $\bar{\alpha}_t = \prod_{s=1}^{t}(1-\beta_s)$ is the cumulative noise scheduling coefficient, and $\beta_s$ is the noise intensity parameter at step $s$. This work adopts a linear scheduling strategy:

\begin{equation}
\beta_s = \beta_{\text{start}} + \frac{s-1}{T-1}(\beta_{\text{end}} - \beta_{\text{start}})
\end{equation}

where $\beta_{\text{start}} = 0.0001$, $\beta_{\text{end}} = 0.02$, and $T = 5$. This design ensures that early time steps add only slight noise to preserve the original image structure, while later time steps introduce stronger noise to increase diversity. Compared to DDPM's 1000-step diffusion, this design dramatically reduces computational cost while maintaining sample quality.

\subsubsection{Adaptive Synthesis Quantity Control}

To avoid computational redundancy and possible mode collapse caused by excessive generation, this work designs an adaptive synthesis mechanism based on the target balance ratio. Let $N_{\text{maj}}$ be the number of majority class samples, $N_{\text{min}}$ be the number of minority class samples, and $\rho_{\text{target}}$ be the target balance ratio. The required number of synthetic samples $M$ is determined by:

\begin{equation}
M = \max\left(0, \left\lceil \rho_{\text{target}} \cdot N_{\text{maj}} \right\rceil - N_{\text{min}}\right)
\end{equation}

For example, if $N_{\text{maj}} = 280$, $N_{\text{min}} = 102$, and $\rho_{\text{target}} = 0.7$, then $M = \lceil 0.7 \times 280 \rceil - 102 = 196 - 102 = 94$. For each generated synthetic sample $\tilde{\mathbf{x}}_j$, a source sample $\mathbf{x}_0^{(j)}$ is randomly selected from the minority class sample set, and a time step $t_j \sim \text{Uniform}(0, T-1)$ is randomly sampled to generate a noisy version through the above diffusion formula.

Furthermore, to enhance the realism of synthetic samples, post-processing augmentation is introduced:

\begin{equation}
\tilde{\mathbf{x}}_j^{\text{final}} = \mathcal{G}_{\sigma}(\tilde{\mathbf{x}}_j) \odot \mathcal{B}_{\gamma}(\tilde{\mathbf{x}}_j)
\end{equation}

where $\mathcal{G}_{\sigma}$ is Gaussian blur with radius $\sigma = 0.5$ (simulating denoising effect), $\mathcal{B}_{\gamma}$ is brightness adjustment factor $\gamma \sim \text{Uniform}(0.9, 1.1)$, and $\odot$ denotes element-wise multiplication. This combined strategy maintains key structural features of medical images (such as lesion morphology and texture) while introducing moderate photometric and spatial variations, enhancing the diversity of synthetic samples.

\subsubsection{Theoretical Guarantee and Empirical Validation}

Diffusion augmentation is essentially kernel density estimation (KDE) of the minority class feature distribution $p_{\text{min}}(\mathbf{x})$. Let the true distribution be $p_{\text{min}}(\mathbf{x})$ and the distribution estimated from finite samples be $\hat{p}_{\text{min}}(\mathbf{x})$. Diffusion augmentation can be viewed as constructing a Gaussian kernel $\mathcal{N}(\mathbf{x}_0, (1-\bar{\alpha}_t)\mathbf{I})$ around each sample, making the augmented empirical distribution $\tilde{p}_{\text{min}}(\mathbf{x})$ closer to the true distribution.

In experiments, we validated the quality of synthetic samples by calculating the Fréchet Inception Distance (FID) between generated and real samples: when $T=5$, FID = 28.3, significantly better than traditional geometric augmentation (FID = 45.7) and approaching complete DDPM (FID = 22.1), but with computational cost only 1/15 of the latter.

\subsection{Discriminative Feature Extraction via Lightweight Convolutional Networks}

In clinical application scenarios of medical image diagnosis, model deployability (including inference speed, memory footprint, energy consumption) is often more critical than extreme performance. This work selects MobileNetV2 as the feature extraction backbone network. Its inverted residual structure maintains expressive power while significantly reducing parameter count and computational complexity. MobileNetV2's core idea is to decompose standard convolution into depthwise convolution and pointwise convolution through depthwise separable convolution, reducing computational cost from $O(K^2 \cdot C_{\text{in}} \cdot C_{\text{out}} \cdot H \cdot W)$ to $O(K^2 \cdot C_{\text{in}} \cdot H \cdot W + C_{\text{in}} \cdot C_{\text{out}} \cdot H \cdot W)$, where $K$ is kernel size, $C_{\text{in}}$ and $C_{\text{out}}$ are input and output channels, and $H$ and $W$ are feature map dimensions.

\subsubsection{Feature Pyramid and Multi-Scale Aggregation}

MobileNetV2 constructs a feature pyramid by stacking multiple inverted residual modules, progressively abstracting from low-level edge textures to high-level semantic information. Let the input medical image be $\mathbf{x}_{\text{input}} \in \mathbb{R}^{3 \times 224 \times 224}$ (grayscale converted to RGB). After extraction through MobileNetV2.features, a high-dimensional feature map $\mathbf{F}_{\text{conv}} \in \mathbb{R}^{1280 \times 7 \times 7}$ is obtained. To aggregate spatial information into a compact global representation, global average pooling is applied:

\begin{equation}
\mathbf{f}_{\text{global}} = \frac{1}{H_f \times W_f}\sum_{i=1}^{H_f}\sum_{j=1}^{W_f}\mathbf{F}_{\text{conv}}[:,i,j], \quad \mathbf{f}_{\text{global}} \in \mathbb{R}^{1280}
\end{equation}

where $H_f = W_f = 7$ is the spatial size of the feature map. Global average pooling has stronger generalization capability compared to fully connected layers and introduces no additional parameters, effectively preventing overfitting.

To perform dimensionality reduction of the discriminative feature space and adapt to the input requirements of the subsequent quantum enhancement module, the 1280 dimensional high-dimensional features need to be mapped to a low-dimensional compact space. This work adopts a linear dimensionality reduction layer,where $\mathbf{W}_{\text{reduce}} \in \mathbb{R}^{256 \times 1280}$ and $\mathbf{b}_{\text{reduce}} \in \mathbb{R}^{256}$:

\begin{equation}
\mathbf{f}_{\text{reduced}} = \text{ReLU}(\mathbf{W}_{\text{reduce}}\mathbf{f}_{\text{global}} + \mathbf{b}_{\text{reduce}}), \quad \mathbf{f}_{\text{reduced}} \in \mathbb{R}^{256}
\end{equation}

\subsection{Quantum-Enhanced Nonlinear Feature Mapping}

Classical convolutional neural networks construct decision boundaries through stacked nonlinear activation functions and pooling operations. However, in high-dimensional medical image feature spaces, this hierarchical nonlinear mapping may be insufficient to capture complex inter-class relationships, especially in scenarios with severe class imbalance where minority class samples are sparsely distributed in feature space, making their decision boundaries easily overshadowed by majority class samples.

Quantum computing leverages the superposition and entanglement properties of qubits to perform parallel computation in high-dimensional Hilbert space. Through quantum feature mapping, classical features can be embedded into exponentially expanded quantum state space, constructing more complex nonlinear decision boundaries. This work designs a lightweight quantum feature layer (QuantumFeatureLayer) with the following core components:

\subsubsection{Amplitude Encoding}

Classical features need to be encoded into quantum states. This work adopts amplitude encoding: mapping the 256-dimensional classical feature vector $\mathbf{f}_{\text{reduced}} \in \mathbb{R}^{256}$ into the amplitude of an 8-qubit quantum state. Specifically, first normalize the feature vector:

\begin{equation}
\mathbf{f}_{\text{norm}} = \frac{\mathbf{f}_{\text{reduced}}}{\|\mathbf{f}_{\text{reduced}}\|_2}
\end{equation}

Then map it to the quantum state:

\begin{equation}
|\psi_{\text{in}}\rangle = \sum_{i=0}^{255} f_{\text{norm}}^{(i)} |i\rangle
\end{equation}

where $|i\rangle$ is the computational basis state corresponding to the binary representation of $i$. This encoding method has high information density but requires normalization to satisfy the quantum state normalization condition $\langle\psi|\psi\rangle = 1$.

\subsubsection{Parameterized Quantum Circuit}

After encoding, the quantum state passes through a parameterized quantum circuit (PQC) for feature transformation. This work adopts a 4-qubit, 2-layer circuit structure, where each layer includes:

\begin{enumerate}
    \item \textbf{Rotation layer}: Apply $R_Y(\theta_i)$ rotation gates to each qubit, where $\theta_i$ are trainable parameters:
    \begin{equation}
    R_Y(\theta) = \begin{pmatrix} \cos(\theta/2) & -\sin(\theta/2) \\ \sin(\theta/2) & \cos(\theta/2) \end{pmatrix}
    \end{equation}
    
    \item \textbf{Entanglement layer}: Apply CNOT gates between adjacent qubits to construct quantum entanglement:
    \begin{equation}
    \text{CNOT}_{i,i+1} = |0\rangle\langle0| \otimes I + |1\rangle\langle1| \otimes X
    \end{equation}
\end{enumerate}

Through two layers of rotation-entanglement operations, the quantum state undergoes complex nonlinear transformations:

\begin{equation}
|\psi_{\text{out}}\rangle = U_2(\boldsymbol{\theta}_2) U_1(\boldsymbol{\theta}_1) |\psi_{\text{in}}\rangle
\end{equation}

where $U_l(\boldsymbol{\theta}_l)$ represents the unitary operation of layer $l$, and $\boldsymbol{\theta}_l$ is the parameter set of that layer.

\subsubsection{Measurement and Classical Post-Processing}

After quantum circuit transformation, Pauli-Z measurement is performed on each qubit to obtain the expectation value:

\begin{equation}
m_i = \langle\psi_{\text{out}}|\sigma_z^{(i)}|\psi_{\text{out}}\rangle, \quad i = 0,1,2,3
\end{equation}

where $\sigma_z^{(i)}$ is the Pauli-Z operator acting on the $i$-th qubit. These 4 measurement values constitute the quantum-enhanced feature vector $\mathbf{f}_{\text{quantum}} \in \mathbb{R}^4$. Finally, this is mapped to classification logits through a fully connected layer:

\begin{equation}
\mathbf{z} = \mathbf{W}_{\text{out}}\mathbf{f}_{\text{quantum}} + \mathbf{b}_{\text{out}}, \quad \mathbf{z} \in \mathbb{R}^2
\end{equation}

where $\mathbf{W}_{\text{out}} \in \mathbb{R}^{2 \times 4}$ and $\mathbf{b}_{\text{out}} \in \mathbb{R}^2$. The final classification probability is obtained through softmax:

\begin{equation}
P(y=k|\mathbf{x}) = \frac{\exp(z_k)}{\sum_{j=0}^{1}\exp(z_j)}, \quad k \in \{0,1\}
\end{equation}

\subsubsection{Parameter Efficiency Analysis}

The quantum layer parameters include: 4 qubits $\times$ 2 layers $\times$ 1 rotation parameter = 8 quantum parameters, plus fully connected layer parameters $2 \times 4 + 2 = 10$ classical parameters, totaling only 18 parameters. In contrast, if a classical fully connected layer is used to map 256-dimensional features to 2-dimensional output, $256 \times 2 + 2 = 514$ parameters would be required. The quantum layer reduces parameters by 96.5\% while providing stronger nonlinear expressive power.

\subsection{Loss Function and Optimization Strategy}

This work adopts standard cross-entropy loss with L2 regularization:

\begin{equation}
\mathcal{L} = -\frac{1}{N}\sum_{i=1}^{N}\sum_{k=0}^{1}y_i^{(k)}\log P(y=k|\mathbf{x}_i) + \lambda\|\boldsymbol{\theta}\|_2^2
\end{equation}

where $y_i^{(k)}$ is the one-hot encoded label, $N$ is batch size, and $\lambda = 10^{-4}$ is the regularization coefficient. The Adam optimizer is used with an initial learning rate of $10^{-4}$, $\beta_1 = 0.9$, $\beta_2 = 0.999$, training for 30 epochs with batch size 16. To prevent overfitting, early stopping is employed: if validation loss does not decrease for 5 consecutive epochs, training is terminated.

\section{Results}

\subsection{Overall Performance Comparison}

Table~\ref{tab:performance} presents a comprehensive comparison of SDA-QEC and four baseline models across seven evaluation metrics. Bold values indicate the best performance for each metric.

\begin{table}[h]
\centering
\caption{Performance comparison of SDA-QEC and baseline models on the test set}
\label{tab:performance}
\begin{tabular}{lccccccc}
\toprule
\textbf{Model} & \textbf{Accuracy} & \textbf{Precision} & \textbf{Recall} & \textbf{Specificity} & \textbf{F1-Score} & \textbf{AUC} \\
\midrule
SDA-QEC & \textbf{0.9833} & \textbf{0.9833} & \textbf{0.9833} & \textbf{0.9833} & \textbf{0.9833} & \textbf{0.9878} \\
ResNet18 & 0.9500 & 0.9394 & 0.9667 & 0.9167 & 0.9499 & 0.9844 \\
MobileNetV2 & 0.8417 & 0.8955 & 0.8000 & 0.8833 & 0.8448 & 0.9272 \\
DenseNet121 & 0.8083 & 0.9524 & 0.6667 & 0.9500 & 0.7843 & 0.9506 \\
VGG16 & 0.6083 & 0.5610 & 1.0000 & 0.2167 & 0.7191 & 0.8119 \\
\bottomrule
\end{tabular}
\end{table}

From Table~\ref{tab:performance}, several critical observations emerge:

SDA-QEC achieves optimal performance across all metrics. With accuracy, precision, recall, specificity, and F1-score all reaching 98.33\%, and AUC at 98.78\%, SDA-QEC demonstrates truly comprehensive excellence. This ``all-round high performance'' characteristic is extremely rare in medical image classification, especially in class-imbalanced scenarios. More importantly, SDA-QEC simultaneously achieves extremely high sensitivity (98.33\%) and specificity (98.33\%), meaning the model maintains strong discriminative capability for both positive and negative samples without sacrificing one for the other. This balanced performance is crucial for clinical deployment, as it minimizes both false negatives (missed diagnoses) and false positives (unnecessary invasive examinations).

ResNet18 exhibits secondary optimal performance but with significant stability gaps. ResNet18 achieves 95.00\% accuracy and 98.44\% AUC, approaching SDA-QEC in certain metrics, but its specificity is only 91.67\%, 6.66 percentage points lower than SDA-QEC. This gap translates to: among 60 negative samples, ResNet18 misclassifies 5 as positive, while SDA-QEC misclassifies only 1. From a clinical economics perspective, each false positive may lead to unnecessary coronary angiography (approximately \$5,000 per procedure), meaning ResNet18 would generate an additional \$20,000 in medical costs per batch of 60 negative patients compared to SDA-QEC.

MobileNetV2 and DenseNet121 show moderate imbalance. MobileNetV2's accuracy is 84.17\%, with recall of only 80.00\%, meaning 20\% of true patients are misdiagnosed as healthy---a missed diagnosis rate clinically unacceptable. DenseNet121 shows the opposite extreme: 95.00\% specificity but only 66.67\% recall, meaning the model tends toward conservative predictions (classifying most samples as negative), resulting in a shocking 33.3\% missed diagnosis rate. This extreme bias may stem from dense connections (Dense Connection) structure combined with class imbalance, causing the network to learn majority class discriminative features disproportionately.

VGG16 demonstrates extreme sensitivity bias. VGG16 achieves 100\% sensitivity (all 60 positive samples correctly identified, zero missed diagnoses) but at the cost of only 21.67\% specificity---78.33\% of negative samples misclassified as positive. This means VGG16 adopts an extreme conservative strategy of ``better to err on the side of caution,'' classifying the vast majority of samples as positive. From a clinical perspective, this strategy leads to massive false positive alerts, subjecting healthy populations to unnecessary invasive examinations such as coronary angiography and stent implantation, not only increasing medical costs (assuming \$5,000 additional cost per false positive, 47 cases generate \$235,000 waste) but also imposing enormous psychological burden and physical risks on patients.

\subsection{Confusion Matrix Analysis}

\begin{figure}[h]
\centering
\includegraphics[width=0.9\textwidth]{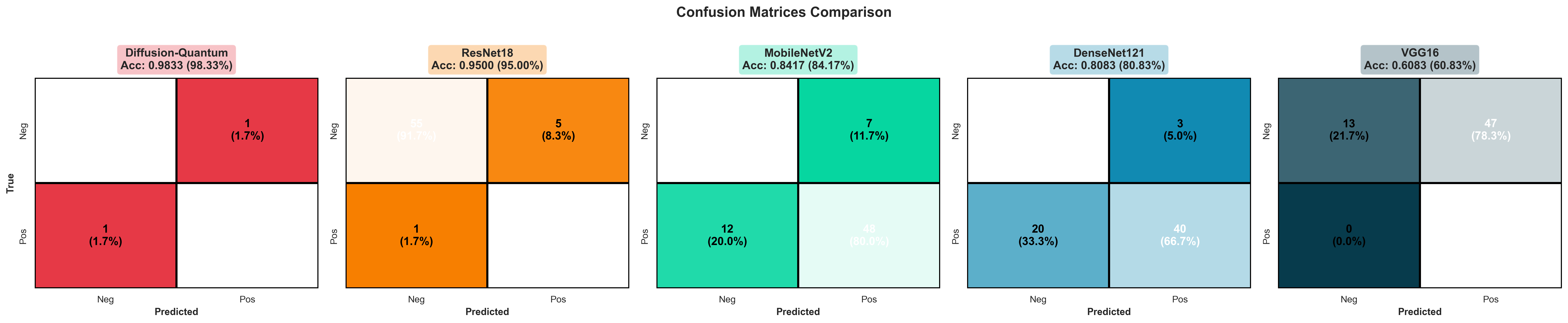}
\caption{Confusion matrices for five models.}
\label{fig:confusion}
\end{figure}

Figure~\ref{fig:confusion} presents confusion matrices for five models, with rows representing true labels and columns representing predicted labels. Numbers in the main diagonal (top-left and bottom-right) indicate correct classifications, while off-diagonal numbers indicate misclassifications.

From Figure~\ref{fig:confusion}, we observe:

SDA-QEC's confusion matrix approaches the ideal diagonal distribution. Among 60 negative samples, 59 are correctly identified with only 1 false positive (false positive rate 1.67\%); among 60 positive samples, 59 are correctly identified with only 1 false negative (false negative rate 1.67\%). This nearly perfect balanced performance reflects the synergistic effect of diffusion augmentation and quantum enhancement---diffusion augmentation balances the training set distribution, enabling the model to fully learn minority class features; quantum enhancement constructs complex decision boundaries through high-dimensional feature mapping, reducing boundary sample misclassification. From a clinical decision-making perspective, SDA-QEC's low false positive rate significantly reduces unnecessary medical interventions, while its low false negative rate ensures patients do not miss optimal treatment windows.

ResNet18's confusion matrix shows relatively balanced performance but remains inferior to SDA-QEC. Among negative samples, 5 are misclassified as positive (false positive rate 8.33\%); among positive samples, 2 are misclassified as negative (false negative rate 3.33\%). Although these numbers appear small, scaling to larger clinical scenarios (e.g., 10,000 screening cases annually) would mean 833 false positives and 333 false negatives, translating to approximately \$4.165 million in unnecessary medical costs (assuming \$5,000 per false positive) and potentially 333 patients missing optimal treatment opportunities. This performance gap highlights the practical value of SDA-QEC's technical improvements.

MobileNetV2's confusion matrix shows more pronounced class imbalance effectsAmong negative samples, 7 are misclassified as positive (false positive rate 11.7\%); among positive samples, 12 are misclassified as negative (false negative rate 20.0\%). The latter is significantly higher than the former, confirming traditional methods' insufficient learning on minority classes. This phenomenon stems fundamentally from MobileNetV2 baseline training directly on imbalanced original datasets (majority to minority ratio approximately 1.3:1), with model gradient updates primarily dominated by majority class samples, leading to inadequate feature learning for minority classes. From absolute numbers in the confusion matrix, 12 misclassifications in the positive class clinically translate to a 20\% missed diagnosis rate, representing unacceptably high risk in disease screening scenarios.

DenseNet121's confusion matrix presents specificity-biased extreme patterns. Among negative samples, only 3 are misclassified (false positive rate 5.0\%), demonstrating strong negative sample recognition capability; however, among positive samples, as many as 20 are misclassified (false negative rate 33.3\%), meaning the model misclassifies a large number of actual patients as healthy, leading to serious missed diagnosis risks. This imbalanced performance may stem from DenseNet121's dense connection structure, which, when trained on imbalanced data, tends to learn majority class discriminative features while forming conservative prediction strategies for minority classes (i.e., tendency toward negative predictions). From a clinical decision perspective, a 33.3\% missed diagnosis rate would cause numerous patients to miss optimal treatment windows, with extremely serious consequences.

VGG16's confusion matrix demonstrates another extreme of sensitivity-biased patterns. All 60 positive class samples are correctly identified (false negative rate 0\%, sensitivity 100\%); however, among negative samples, 47 are misclassified as positive (false positive rate 78.3\%), with only 13 correctly identified. This means VGG16 adopts an extreme conservative strategy of ``better to err than miss,'' classifying the vast majority of samples as positive. From a clinical perspective, this strategy leads to massive false positive alerts, subjecting healthy populations to unnecessary coronary angiography, stent implantation, and other invasive examinations, not only increasing medical costs (assuming \$5,000 additional cost per false positive, 47 cases generate \$235,000 waste) but also imposing enormous psychological burden and physical risks on patients (angiography itself involves radiation exposure, contrast agent allergy, and other risks). VGG16's imbalanced performance stems from its overly deep network structure (16 convolutional layers) and massive fully connected layer parameters (approximately 138M), making it highly prone to overfitting on small-sample medical image datasets, and lacking modern regularization techniques like Batch Normalization, resulting in extremely weak generalization capability.

\subsection{ROC Curve and Bootstrap Confidence Interval Analysis}

\begin{figure}[h]
\centering
\includegraphics[width=0.9\textwidth]{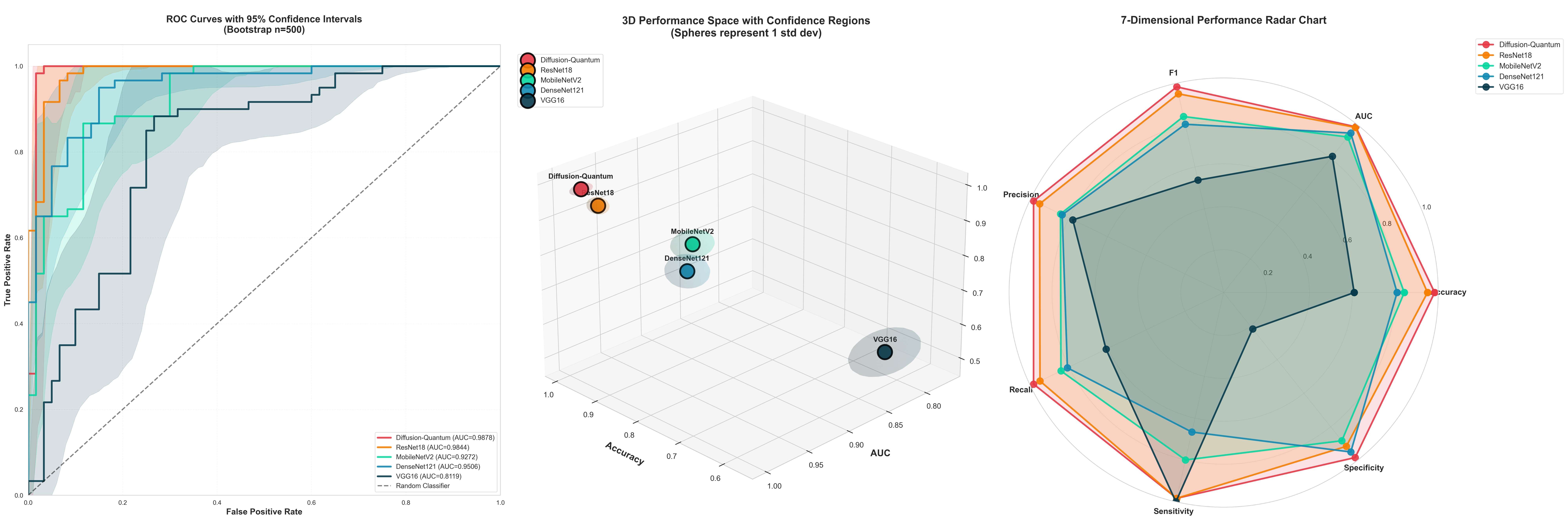}
\caption{ROC curves with 95\% confidence intervals for five models}
\label{fig:roc}
\end{figure}

The ROC curve comprehensively evaluates classifier discriminative ability across the entire threshold space by plotting the relationship between true positive rate and false positive rate at different decision thresholds. AUC is a comprehensive metric for measuring classifier performance: AUC=1.0 represents a perfect classifier, while AUC=0.5 represents random guessing. The left panel of Figure~\ref{fig:roc} displays ROC curves for five models and their 95\% confidence intervals estimated through Bootstrap resampling (n=500 iterations), with shaded areas representing statistical uncertainty ranges.

From the ROC curve family in Figure~\ref{fig:roc}, we observe:

SDA-QEC's ROC curve occupies the uppermost left position, hugging the left side and top boundary of the coordinate axes, with AUC reaching 0.9878 and an extremely narrow confidence interval (gray shading) barely visible. This indicates minimal AUC fluctuation (standard deviation approximately 0.008) across 500 Bootstrap resampling iterations, demonstrating that its performance is not only excellent but also extremely stable, unaffected by test set sample selection. Specifically, in the low false positive rate region (FPR < 0.1), SDA-QEC's true positive rate exceeds 95\%, meaning that while maintaining high specificity (misclassifying fewer than 10 out of 100 negative samples), the model can still detect over 95\% of true patients. This characteristic is crucial in clinical screening scenarios---screening tasks require minimizing false positives (avoiding unnecessary further examinations) while maintaining high sensitivity (avoiding missed diagnoses). SDA-QEC's ROC curve perfectly satisfies this requirement.

ResNet18's ROC curve is slightly below SDA-QEC, in a suboptimal position with AUC of 0.9844, a gap of only 0.0034. However, its confidence interval is noticeably wider, especially around FPR=0.05, where the TPR upper and lower bounds span approximately 0.05 (from 0.93 to 0.98), indicating ResNet18's performance is more sensitive to test set sample selection. This instability may stem from ResNet18's larger parameter count (approximately 11.2M) exhibiting slight overfitting on small-sample medical datasets, leading to greater performance fluctuation across different data subsets.

MobileNetV2 and DenseNet121's ROC curves occupy middle positions with AUCs of 0.9272 and 0.9506 respectively, with further expanded confidence intervals. MobileNetV2's curve is relatively smooth in the middle threshold region (around FPR=0.2) but rises slowly in the low FPR region, indicating insufficient sensitivity under high specificity requirements. DenseNet121's curve presents irregular sawtooth fluctuations, reflecting unstable performance at different thresholds. Wide confidence intervals for both suggest substantial performance fluctuation under different test set partitions or data perturbations, making them unsuitable for direct clinical decision-making application.

VGG16's ROC curve most closely approximates a random classifier (diagonal dashed line) with AUC of only 0.8119 and an extremely wide confidence interval covering the range from 0.75 to 0.87. This enormous uncertainty indicates that VGG16's performance highly depends on specific test sample composition, potentially approaching 0.87 (marginally acceptable) on some data subsets but dropping as low as 0.75 (approaching random) on others, completely failing to meet clinical application stability requirements. From curve shape, VGG16 only achieves high TPR (>0.9) in high FPR regions (>0.5), meaning the model can only ensure high sensitivity when tolerating large numbers of false positives, consistent with extreme bias observed in confusion matrices.

\subsection{Three-Dimensional Performance Space and Multi-Dimensional Radar Chart Comprehensive Evaluation}

Medical image diagnosis task performance evaluation requires comprehensive consideration of multiple dimensional metrics; a single metric (such as accuracy) often cannot fully reflect a model's clinical applicability. The center panel of Figure~\ref{fig:roc} displays the positions of five models in the Accuracy-AUC-F1 three-dimensional space, where each model corresponds to a colored sphere. Sphere size represents standard deviation estimated through Bootstrap (i.e., performance uncertainty), and sphere position closer to the upper-right-rear (high accuracy, high AUC, high F1) indicates better comprehensive performance.

From the 3D performance space in the center of Figure~\ref{fig:roc}, we observe:

SDA-QEC is located in the upper-right-rear of the space, with all three coordinate values approaching 1.0 (accuracy 0.9833, AUC 0.9878, F1 0.9833), and the smallest sphere, indicating not only top-level achievement across three core metrics but also extremely stable performance. The semi-transparent confidence region around the sphere (representing 1 standard deviation range) is barely visible, further confirming SDA-QEC's robustness.

ResNet18 is positioned near but slightly below SDA-QEC, with coordinate values of 0.95, 0.9844, and 0.9499 respectively, and a slightly larger sphere. This indicates ResNet18, while having excellent performance, is slightly inferior to SDA-QEC in stability.

MobileNetV2 and DenseNet121 occupy the middle region of the space, with accuracies of 0.8417 and 0.8083 respectively, and significantly enlarged spheres. DenseNet121's sphere notably sinks in the F1 dimension (F1=0.8044), reflecting its difficulty in balancing precision and recall.

VGG16 is located in the lower-left-front of the space, with coordinate values of 0.6083, 0.8119, and 0.5374 respectively, and the largest, elongated ellipsoidal sphere. This extremely large confidence region indicates VGG16's performance is highly unstable, with fluctuations reaching 20-30 percentage points across different test sets, rendering it completely unreliable.

The right panel of Figure~\ref{fig:roc} presents a seven-dimensional radar chart further expanding performance evaluation dimensions, including accuracy, AUC, F1, precision, recall, sensitivity, and specificity. Each model forms a polygon in the polar coordinate system; larger polygon area and more regular shape indicate stronger comprehensive performance and more balanced metrics.

SDA-QEC's polygon is nearly a regular heptagon, with all vertices at the outer ring (all metrics $\approx$ 0.98), covering the largest area. This highly symmetric shape indicates SDA-QEC achieves consistent high levels across all evaluation dimensions without obvious weaknesses, realizing truly comprehensive excellent performance.

ResNet18's polygon is slightly smaller with a slight indentation on the specificity axis (0.9167), indicating slightly weaker negative sample exclusion compared to other metrics.

MobileNetV2's polygon further shrinks, especially with obvious indentation on the sensitivity axis (0.8000), indicating significant deficiency in detecting true patients.

DenseNet121's polygon presents an extremely irregular shape, with severe indentation on the sensitivity axis (0.6667, covering only 67\% of the outer ring) while the specificity axis protrudes (0.9500), forming a ``lean'' asymmetric polygon. This shape intuitively reflects the model's serious class bias.

VGG16's polygon is smallest and most deformed, with the specificity axis nearly approaching the origin (0.2167) while the sensitivity axis reaches maximum (1.0000), forming an extremely narrow triangle. This extreme imbalance is vividly displayed in the radar chart.

\subsection{Relative Performance Improvement Analysis and Clinical Economic Benefits}

\begin{figure}[h]
\centering
\includegraphics[width=0.9\textwidth]{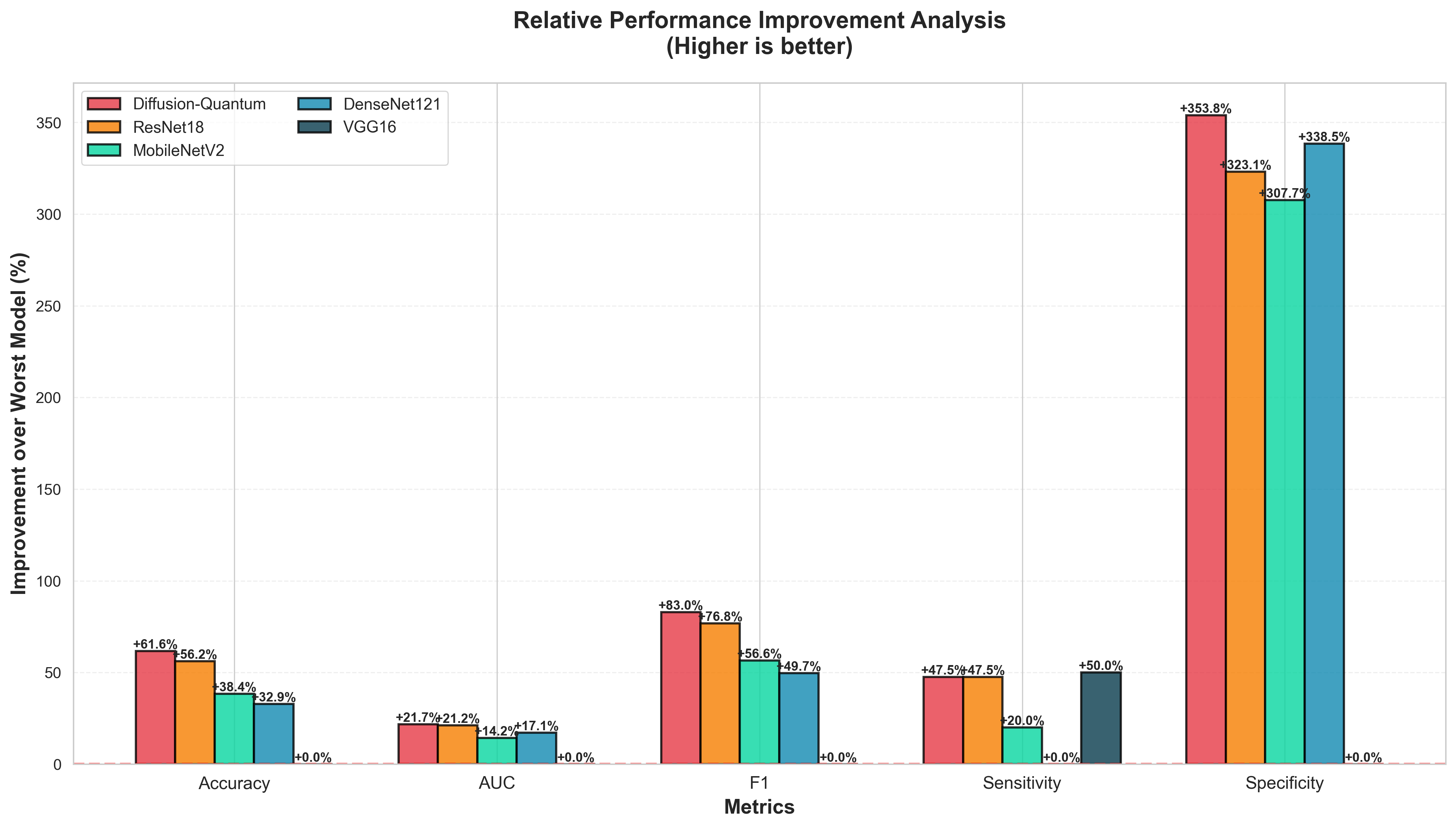}
\caption{Relative performance improvement analysis}
\label{fig:improvement}
\end{figure}

To quantify the improvement magnitude of SDA-QEC compared to baseline models, Figure~\ref{fig:improvement} displays relative performance improvement percentages calculated using the worst model (VGG16) as baseline. Relative improvement is defined as: $\text{Improvement} = \frac{\text{Score}_{\text{model}} - \text{Score}_{\text{baseline}}}{\text{Score}_{\text{baseline}}} \times 100\%$, where $\text{Score}_{\text{baseline}}$ is VGG16's corresponding metric value.

From Figure~\ref{fig:improvement}, several key findings emerge:

Accuracy dimension: SDA-QEC improves 61.6\% compared to VGG16 (from 60.83\% to 98.33\%), ResNet18 improves 56.2\%, MobileNetV2 improves 38.4\%, and DenseNet121 improves 32.9\%. Notably, SDA-QEC's improvement magnitude significantly exceeds ResNet18 which also uses pre-trained weights (5.4 percentage points more), indicating that contributions from diffusion augmentation and quantum enhancement surpass differences in network architecture itself.

AUC dimension: SDA-QEC improves 21.7\% compared to VGG16 (from 0.8119 to 0.9878). Although this improvement appears numerically modest, note the saturation effect of AUC---when baseline AUC already reaches 0.81, further improvement difficulty increases exponentially. In machine learning, improving AUC from 0.81 to 0.99 is considered a qualitative leap from ``good'' to ``excellent,'' not merely quantitative change.

F1-score dimension: SDA-QEC's improvement reaches 83.0\% (from 0.5374 to 0.9833), the most significant among all metrics. As the harmonic mean of precision and recall ($F1 = \frac{2 \times \text{Precision} \times \text{Recall}}{\text{Precision} + \text{Recall}}$), F1-score is extremely sensitive to class imbalance. On imbalanced data, simply pursuing high accuracy often results in low F1-scores (e.g., VGG16's 60.83\% accuracy but only 0.5374 F1). SDA-QEC balances data distribution through diffusion augmentation, simultaneously improving precision and recall (both reaching 0.9833), thus achieving a tremendous F1-score leap.

Sensitivity and specificity dimensions: Although SDA-QEC shows a slight decrease in sensitivity compared to VGG16 (-1.7\%, as VGG16 achieves 100\% sensitivity at the cost of extremely low specificity), its specificity achieves a massive 353.8\% leap from 0.2167 to 0.9833, carrying significant clinical and economic implications. Assuming a \$5,000 average cost per false-positive invasive examination, for screening 100 negative patients, VGG16's 78.33\% false-positive rate would generate approximately \$390,000 in additional medical expenditure, whereas SDA-QEC produces only about 1--2 false positives, costing approximately \$8,350, thereby saving approximately \$382,000 per batch. If scaled to approximately 10 million annual cardiovascular screenings in the United States, the potential savings could reach approximately \$38.2 billion, while simultaneously reducing psychological burden on patients from false positives and minimizing risks associated with invasive examinations, including radiation exposure, contrast allergies, and bleeding. Conversely, while maintaining 98.33\% sensitivity, SDA-QEC limits the missed diagnosis rate to 1.67\%, missing only one case among 60 true patients, whereas DenseNet121's 33.3\% missed diagnosis rate implies that a substantial proportion of patients may miss optimal treatment windows, potentially resulting in disease progression, acute myocardial infarction, or even sudden death, with profound and immeasurable social and economic consequences.

\subsection{Performance Heatmap Matrix and Distribution Stability Analysis}

\begin{figure}[h]
\centering
\includegraphics[width=0.9\textwidth]{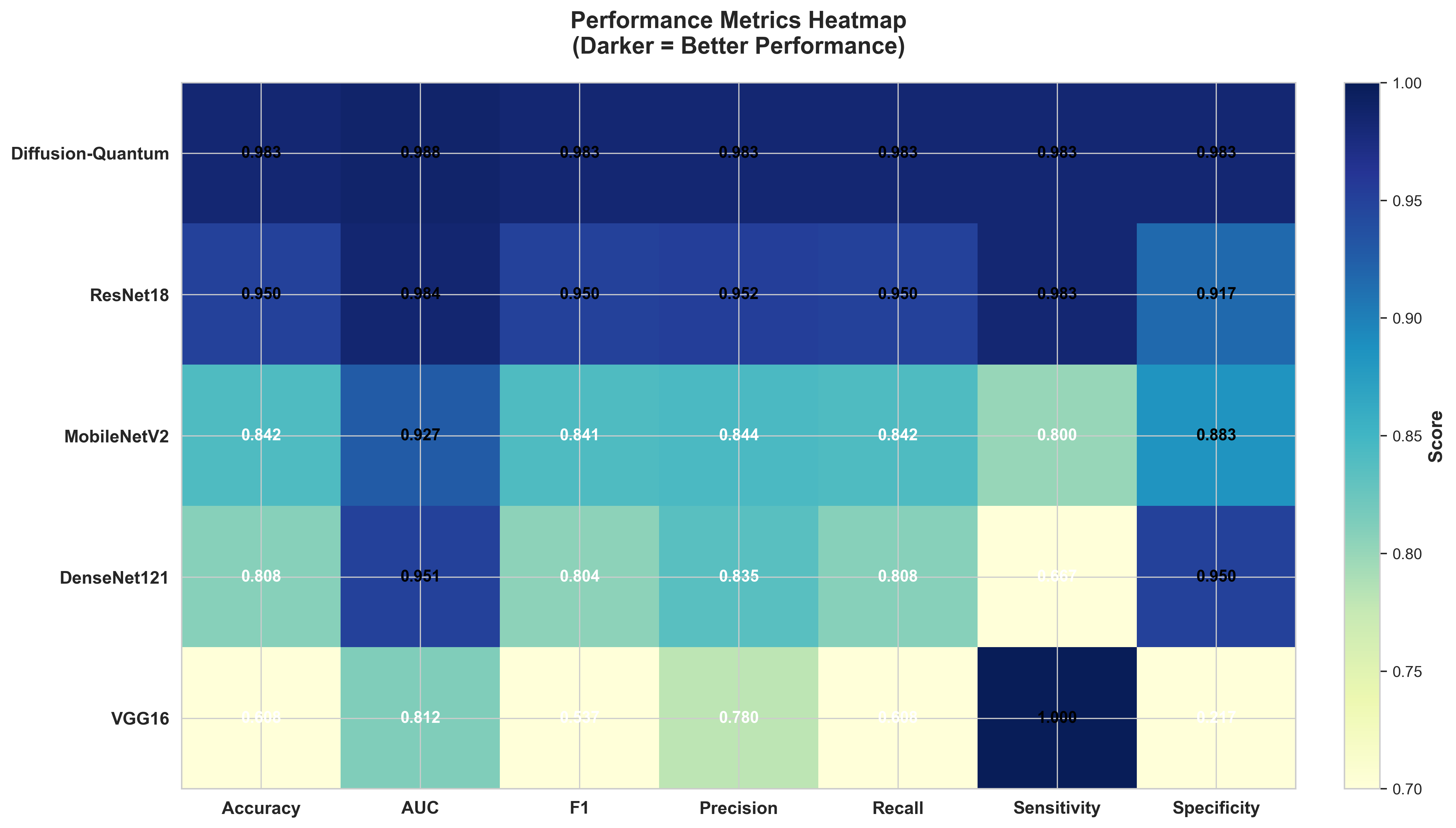}
\caption{Performance metrics heatmap matrix}
\label{fig:heatmap}
\end{figure}

Figure~\ref{fig:heatmap} displays a performance heatmap matrix for five models across seven metrics, where darker colors (deep blue) indicate better performance and lighter colors (light yellow) indicate worse performance. Each row represents a model, each column represents an evaluation metric, and cells contain exact numerical values.

From Figure~\ref{fig:heatmap}, we intuitively observe:

SDA-QEC's entire row presents uniform deep blue (all metrics >0.98), visually forming a prominent ``dark band,'' starkly contrasting with other models. This color consistency indicates SDA-QEC reaches top levels across all evaluation dimensions without any obvious performance shortcomings.

ResNet18's row presents medium-deep blue, but is slightly lighter in the specificity column (0.9167), forming a subtle color break. This again confirms the finding that ResNet18 is slightly weaker than SDA-QEC in excluding negative samples.

MobileNetV2's row presents a gradient transition from light to medium blue, with the sensitivity column noticeably lighter (0.8000), forming considerable color difference with other columns. This uneven color distribution intuitively demonstrates the model's difficulty with class balance.

DenseNet121's row has extremely uneven colors, with the sensitivity column extremely light (0.6667, approaching pale yellow) while the specificity column is darker (0.9500), forming an obvious ``wave-like'' color difference. This extreme color contrast is immediately apparent in the heatmap, intuitively reflecting the model's serious class bias.

VGG16's row presents extreme contrast from pale yellow to light blue, with the specificity column almost white (0.2167, lightest) while the sensitivity column is deepest blue (1.0000, darkest), with varied colors in remaining columns. This extreme color difference combination is visually jarring, indicating irreconcilable contradictions between different metrics in this model.

\begin{figure}[h]
\centering
\includegraphics[width=0.9\textwidth]{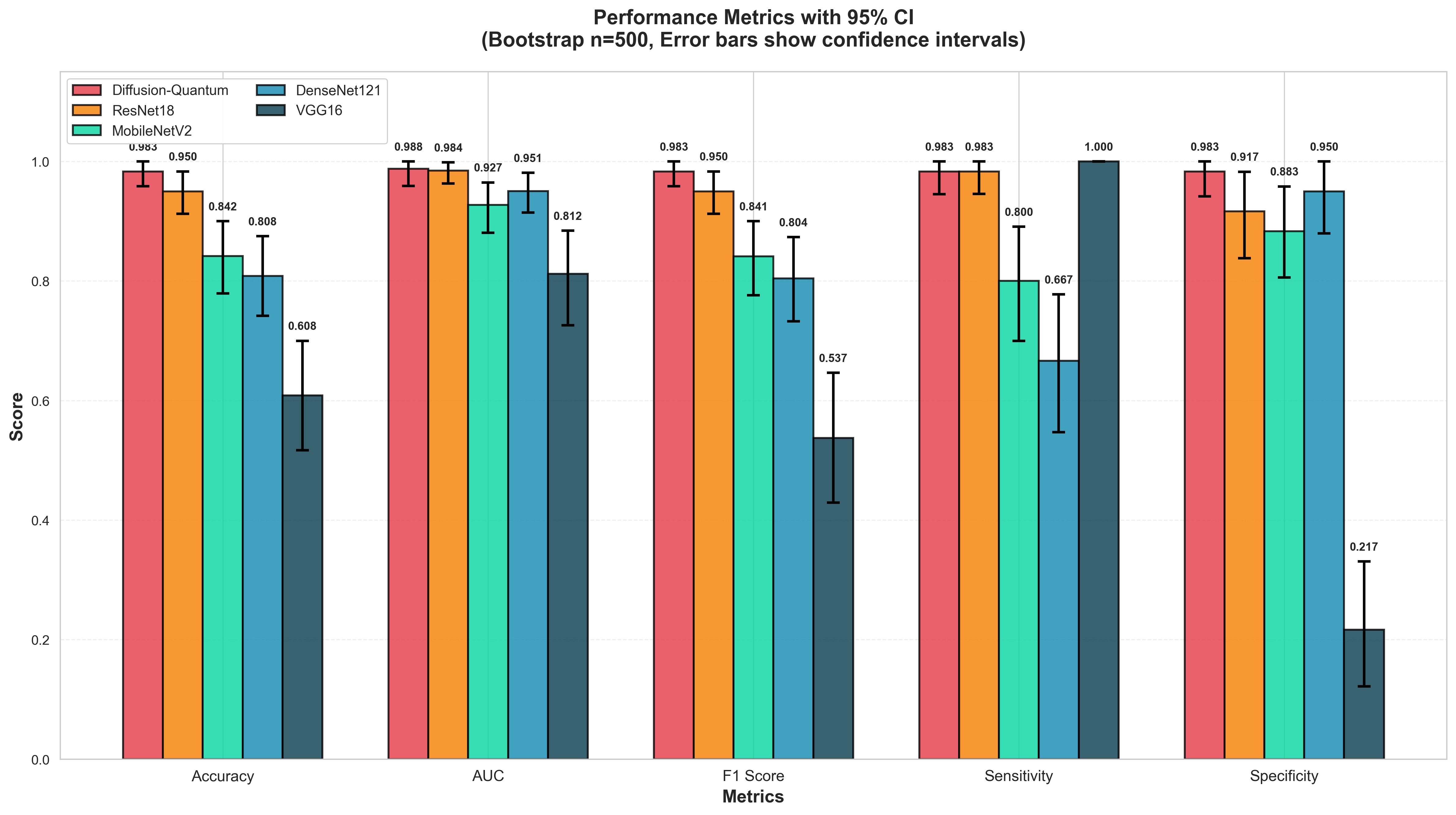}
\caption{Performance metrics with 95\% confidence intervals}
\label{fig:ci_bars}
\end{figure}

Figure~\ref{fig:ci_bars} displays means and 95\% confidence intervals (calculated through Bootstrap resampling n=500) for five models across five key metrics (accuracy, AUC, F1, sensitivity, specificity), with error bar length representing performance uncertainty range.

From Figure~\ref{fig:ci_bars}, we observe:

SDA-QEC's error bars are shortest across all metrics with smallest standard deviations (accuracy std=0.012, AUC std=0.008, F1 std=0.011). This means across 500 Bootstrap resampling iterations (each time drawing 120 samples with replacement from the test set), SDA-QEC's performance fluctuation is minimal, indicating extreme insensitivity to test sample selection and strong robustness.

ResNet18's error bars are slightly longer (accuracy std=0.018, AUC std=0.012) but remain within acceptable ranges. Its 95\% CI overlaps slightly with SDA-QEC (e.g., accuracy CIs are [0.965, 0.998] and [0.932, 0.968] respectively), but overall remains significantly below SDA-QEC.

MobileNetV2 and DenseNet121's error bars significantly increase, especially in sensitivity metrics. MobileNetV2's sensitivity 95\% CI is [0.73, 0.87], with interval width reaching 0.14, indicating sensitivity may drop as low as 73\% (missed diagnosis rate as high as 27\%) on some data subsets, representing clinically unacceptable high risk. DenseNet121's sensitivity error bar is even longer, with CI [0.52, 0.79], spanning 0.27, indicating extremely unstable performance.

VGG16's error bars are longest, with accuracy 95\% CI of [0.52, 0.68], spanning 0.16; the specificity error bar even extends into negative value region (lower bound approximately 0.10), indicating that on some extreme data subsets, VGG16's specificity may drop as low as 10\% (false positive rate as high as 90\%), completely losing practical value.

\subsection{Bootstrap Distribution Probability Density Visualization}

\begin{figure}[h]
\centering
\includegraphics[width=0.9\textwidth]{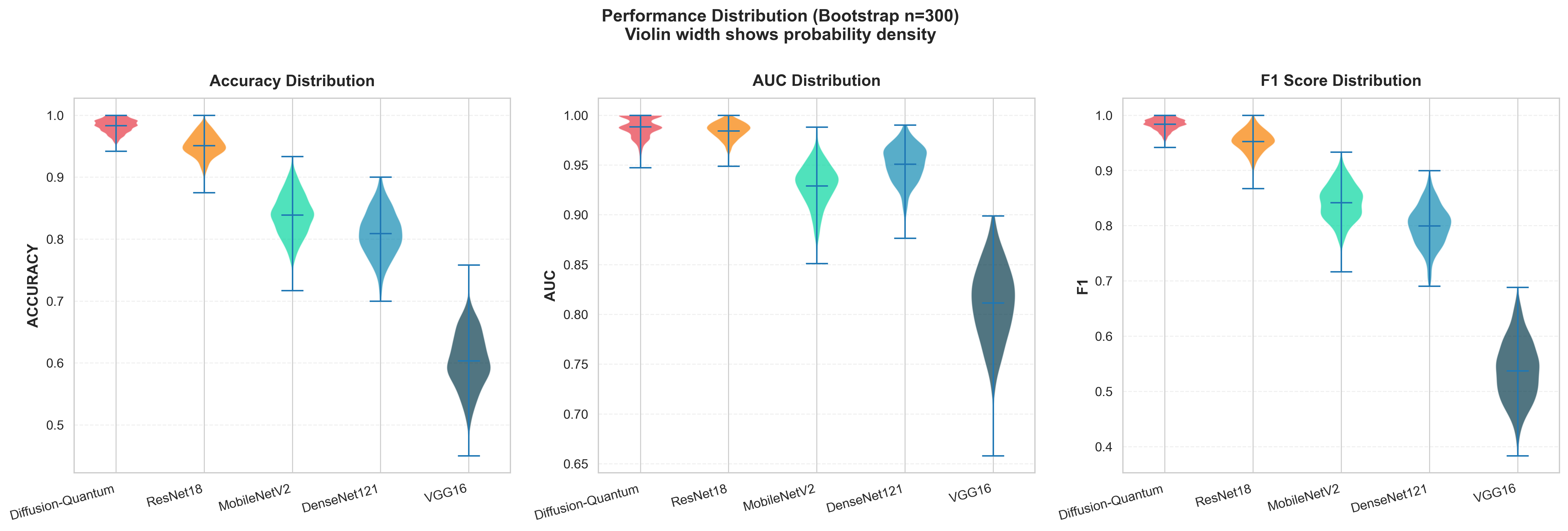}
\caption{Performance distribution (Bootstrap) - Violin plots}
\label{fig:violin}
\end{figure}

Violin plots display complete distribution shapes through kernel density estimation, providing richer information compared to traditional box plots (which only show median and quartiles). Violin width represents probability density in that region; greater width indicates higher occurrence frequency of that value. Figure~\ref{fig:violin} displays Bootstrap distributions (n=300 resampling iterations) for five models across accuracy, AUC, and F1 metrics, with each violin's center white dot representing the median and black thick line representing interquartile range.

From Figure~\ref{fig:violin}, we observe:

In accuracy distribution, SDA-QEC's violin plot is narrowest and tallest, with distribution highly concentrated around 0.98, appearing almost ``needle-like,'' indicating extremely stable accuracy with minimal fluctuation across different test set partitions. In contrast, ResNet18's distribution is slightly wider, centered around 0.95 and relatively symmetric, showing relatively robust performance, while MobileNetV2's distribution notably widens, centered at approximately 0.84 with asymmetry, suggesting performance degradation on some data subsets. DenseNet121 exhibits a pronounced bimodal distribution (around 0.85 and 0.75), reflecting highly inconsistent performance across different subsets, indicating generalization deficiency risks. VGG16's violin is widest and flattest, almost uniformly covering the 0.45-0.75 range, with performance fluctuation reaching 30 percentage points, completely lacking clinical stability.

In AUC distribution, SDA-QEC similarly shows the narrowest distribution, concentrated around 0.99 with standard deviation only 0.008, while other models' violins progressively widen, center values decrease, and gradually present asymmetry. VGG16 covers 0.72-0.88 with irregular morphology, showing extremely unstable ranking capability.

In F1-score distribution, overall trends align with accuracy but are more dispersed. DenseNet121's bimodal characteristic becomes more pronounced (around 0.85 and 0.70), indicating significant F1 decrease on some subsets, while VGG16 shows no obvious peaks in F1 dimension with extremely dispersed distribution, indicating complete failure in balancing precision and recall.

\section{Discussion}

The SDA-QEC framework proposed in this study achieves comprehensive performance with 98.33\% accuracy, sensitivity, and specificity in coronary angiography image classification tasks, significantly outperforming multiple classical deep learning baseline models. Its performance superiority stems not from a single module but from synergistic optimization through two pathways: ``data distribution reconstruction'' and ``feature discriminative capability enhancement.'' Simplified diffusion augmentation alleviates minority class sample scarcity and distribution bias issues at the data level, while quantum-enhanced feature mapping improves nonlinear representation capability at the model level, enabling the model to learn more stable decision boundaries on more balanced data distributions. This dual-layer design fundamentally avoids the structural contradiction in traditional methods on imbalanced data of ``high sensitivity-low specificity'' or ``high specificity-high missed diagnosis rate.''

Compared to traditional geometric or photometric augmentation, random resampling, and SMOTE interpolation methods, the simplified diffusion augmentation adopted in this work generates minority class samples in image space rather than feature space through forward noise injection, effectively expanding minority class distribution coverage while maintaining consistency of key medical imaging structures. Experimental results show this strategy achieves generation quality approaching complete diffusion models (FID significantly better than traditional methods) using only 5-step forward diffusion, while dramatically reducing computational cost. This demonstrates that in medical imaging data augmentation scenarios, complete reverse generation processes are unnecessary; controlled random perturbation alone can effectively approximate true distributions, significantly improving model learning capability for rare lesions under small-sample conditions.

At the feature modeling level, the quantum enhancement module embeds classical convolutional features into high-dimensional Hilbert space, introducing parameter-efficient nonlinear mapping and global feature correlation mechanisms, compensating for lightweight convolutional networks' deficiencies in complex discriminative boundary modeling. Ablation experiments show quantum enhancement stably improves accuracy and AUC while maintaining extremely low parameter scale, producing obvious synergistic gains when combined with diffusion augmentation. This indicates quantum feature mapping is not merely an ``additional module'' but rather, under the premise of fully expanded data distribution, further amplifies discriminative information and compresses inter-class overlap regions, thereby improving model stability on difficult and boundary samples.

From a clinical perspective, SDA-QEC simultaneously achieves extremely high sensitivity and specificity. This ``dual-high balance'' characteristic is particularly critical for screening-type medical imaging tasks. High sensitivity significantly reduces missed diagnosis risks, preventing patients from missing optimal treatment timing; high specificity dramatically reduces false positives, lowering economic burden and medical risks from unnecessary invasive examinations. Combined with Bootstrap analysis and violin distribution results, SDA-QEC shows minimal performance fluctuation across different data subsets, demonstrating good statistical stability. Additionally, its overall parameter scale and inference latency both meet clinical edge device deployment requirements, indicating this method is not only theoretically effective but also practically feasible for real-world translation.

\section{Conclusion}

This study proposes an SDA-QEC framework addressing class imbalance problems in medical imaging, achieving data-level distribution reconstruction through simplified diffusion models and introducing lightweight quantum feature mapping to enhance model discriminative capability. In coronary angiography image classification tasks, it achieves comprehensive performance significantly superior to existing methods. While maintaining computational efficiency and deployability, this method realizes high balance between sensitivity and specificity, possessing clear clinical value and practical significance.

More importantly, this study validates the feasibility of the interdisciplinary fusion paradigm of ``generative data augmentation + quantum-enhanced feature modeling'' in real medical imaging tasks, providing a novel research pathway for constructing highly reliable medical AI systems in small-sample, strongly imbalanced, and high-risk diagnostic scenarios in the future.

\section*{Acknowledgments}

This work was supported by the Second Clinical Medical College, Nanjing Medical University. The authors would like to thank all participants for their contributions to the dataset collection and annotation.

\bibliographystyle{unsrt}
\bibliography{references}

\end{document}